\documentclass{article} 
\usepackage{iclr2020_conference,times}

\usepackage{amsmath,amsfonts,bm}

\def\eqref#1{equation~\ref{#1}}

\def\1{\bm{1}}

\DeclareMathAlphabet{\mathsfit}{\encodingdefault}{\sfdefault}{m}{sl}
\SetMathAlphabet{\mathsfit}{bold}{\encodingdefault}{\sfdefault}{bx}{n}

\usepackage{hyperref}
\usepackage{url}
\usepackage{booktabs}
\usepackage{graphicx}
\usepackage{tabularx}
\usepackage{subcaption}
\usepackage{xargs}
\usepackage{array}
\newcolumntype{S}{>{\centering\arraybackslash}m{1.6cm}}
\newcolumntype{O}{>{\centering\arraybackslash}m{1.2cm}}
\newcolumntype{L}{>{\centering\arraybackslash}m{1.18cm}}
\newcolumntype{D}{>{\centering\arraybackslash}m{1cm}}
\usepackage[colorinlistoftodos,prependcaption,textsize=tiny]{todonotes}
\newcommandx{\zack}[1]{\textcolor{blue}{#1 -Zack}}
\newcommandx{\dk}[1]{\textcolor{red}{#1 -DK}}

\newcommand{\expnumber}[2]{{#1}\mathrm{e}{#2}}

\title{Learning the Difference that Makes a Difference
with Counterfactually-Augmented Data
}

\author{Divyansh Kaushik, Eduard Hovy, Zachary C. Lipton\\ 
Carnegie Mellon University\\
Pittsburgh PA, USA \\
\texttt{\{dkaushik, hovy, zlipton\}@cmu.edu} \\
}

 \iclrfinalcopy 
\begin{document}

\maketitle
\begin{abstract}
Despite alarm over the reliance of machine learning systems
on so-called \emph{spurious} patterns,
the term lacks coherent meaning in standard statistical frameworks.
However, the language of causality offers clarity:
spurious associations are due to confounding (e.g., a common cause),
but not direct or indirect causal effects.
In this paper, we focus on natural language processing, 
introducing methods and resources 
for training models less sensitive to spurious patterns.
Given documents and their initial labels,
we task humans with revising each document
so that it (i) accords with a counterfactual target label;
(ii) retains internal coherence; 
and (iii) avoids unnecessary changes.
Interestingly, on \emph{sentiment analysis} 
and \emph{natural language inference} tasks,
classifiers trained on original data fail on their 
counterfactually-revised counterparts and vice versa.
Classifiers trained on combined datasets 
perform remarkably well,
just shy of those specialized to either domain. 
While classifiers trained on either original or manipulated data alone 
are sensitive to spurious features (e.g., mentions of \emph{genre}),
models trained on the combined data are less sensitive to this signal.
Both datasets are publicly available\footnote{\href{https://github.com/dkaushik96/counterfactually-augmented-data}{https://github.com/dkaushik96/counterfactually-augmented-data}}. 

\end{abstract}

\section{Introduction}
\emph{What makes a document's sentiment positive?
What makes a loan applicant creditworthy? 
What makes a job candidate qualified?
When does a photograph truly depict a dolphin?
Moreover, what does it mean for a feature 
to be relevant to such a determination?}

Statistical learning offers one framework 
for approaching these questions.
First, we swap out the semantic question 
for a more readily answerable associative question. 
For example, instead of asking 
\emph{what conveys a document's sentiment},
we recast the question as 
\emph{which documents are \underline{likely} 
to be labeled as positive (or negative)?}
Then, in this associative framing, 
we interpret as \emph{relevant}, 
those features that are most \emph{predictive} of the label. 
However, despite the rapid adoption 
and undeniable commercial success  
of associative learning, this framing seems unsatisfying.

Alongside deep learning's predictive wins, 
critical questions have piled up 
concerning \emph{spurious patterns}, \emph{artifacts}, 
\emph{robustness}, and \emph{discrimination},
that the purely associative perspective appears ill-equipped to answer.
For example, in computer vision, researchers have found 
that deep neural networks rely on surface-level texture 
\citep{jo2017measuring, geirhos2018imagenet}
or clues in the image's background to recognize foreground objects 
even when that seems both unnecessary and somehow wrong: 
\emph{the beach is not what makes a seagull a seagull}.
And yet, researchers struggle to articulate precisely why
models \emph{should not} rely on such patterns.

In natural language processing (NLP), 
these issues have emerged as central concerns
in the literature on \emph{annotation artifacts}
and \emph{societal biases}.
Across myriad tasks, researchers have demonstrated 
that models tend to rely on \emph{spurious} associations \citep{poliak2018hypothesis,gururangan2018annotation,kaushik2018much,kiritchenko2018examining}.
Notably, some models for question-answering tasks
may not actually be sensitive to the choice of the question \citep{kaushik2018much},
while in \emph{Natural Language Inference} (NLI), 
classifiers trained on \emph{hypotheses} only (vs hypotheses and premises)
perform surprisingly well \citep{poliak2018hypothesis, gururangan2018annotation}.
However, papers seldom make clear what, if anything, \emph{spuriousness} means
within the standard supervised learning framework.
ML systems are trained to exploit the mutual information 
between features and a label to make accurate predictions.
The standard statistical learning toolkit does not offer a conceptual distinction 
between spurious and non-spurious associations. 

\begin{figure}[t]
\centering
    \includegraphics[width=0.7\textwidth]{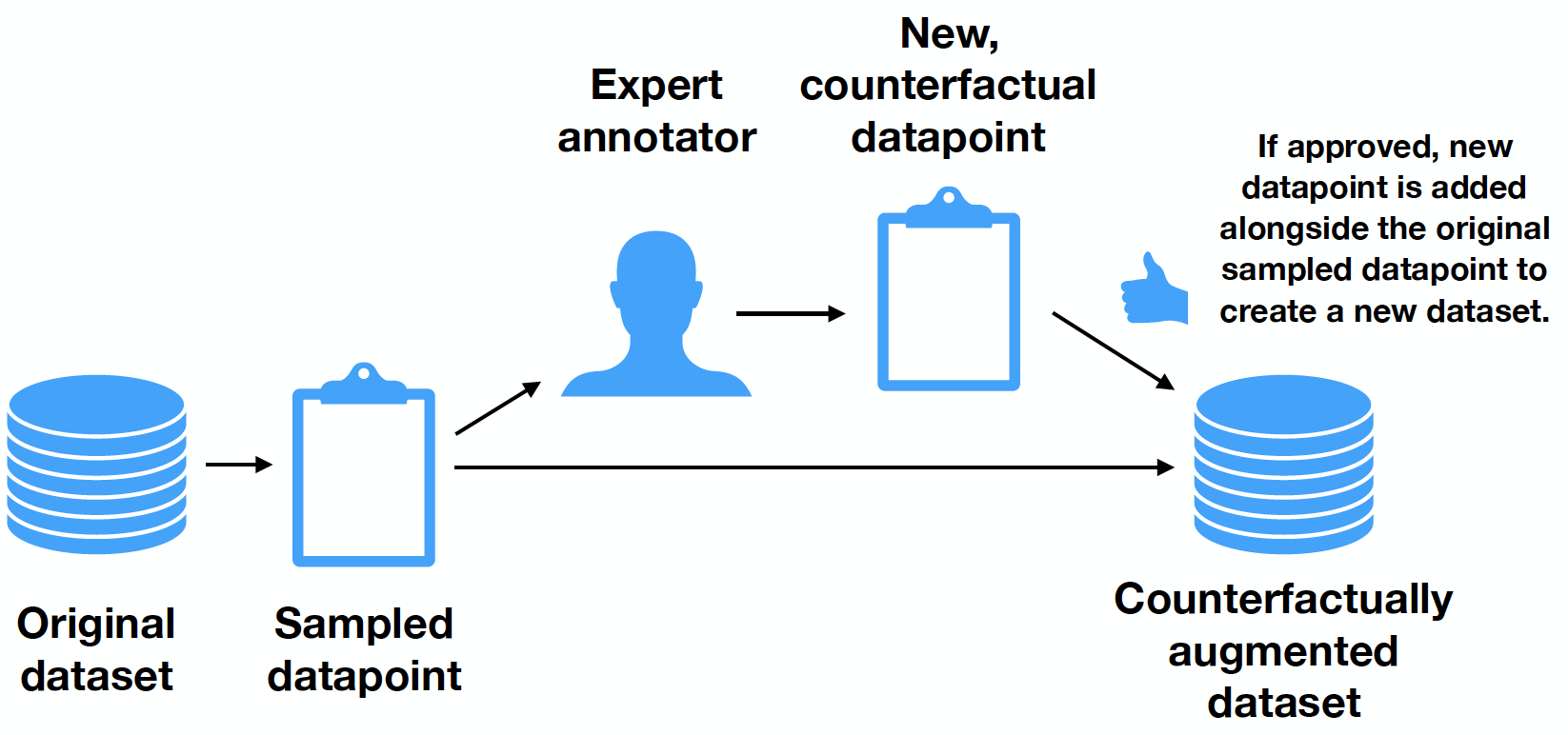}
\caption{
Pipeline for collecting and leveraging counterfactually-altered data
\label{fig:pipeline}}
\vspace{-10px}
\end{figure}

Causality, however, offers a coherent notion of spuriousness.
Spurious associations owe to confounding 
rather than to a (direct or indirect) causal path. 
We might consider a factor of variation 
to be spuriously correlated with a label of interest if intervening upon it
would not impact the applicability of the label or vice versa.
While our paper does not call upon 
the mathematical machinery of causality, 
we draw inspiration from the underlying philosophy
to design a new dataset creation procedure 
in which humans \emph{counterfactually revise} documents.

Returning to NLP, although we lack automated tools 
for mapping between raw text and disentangled factors,
we nevertheless describe documents 
in terms of these abstract representations.
Moreover, it seems natural to speak of manipulating these
factors directly \citep{hovy1987generating}.
Consider, for example, the following interventions:
(i) \emph{Revise the letter to make it more positive};
(ii) \emph{Edit the second sentence so that it appears to contradict the first}.
These edits might be thought of as intervening on only those
aspects of the text that are necessary 
to make the counterfactual label applicable.

In this exploratory paper, we design a human-in-the-loop system 
for counterfactually manipulating documents. 
Our hope is that by intervening only upon the factor of interest,
we might disentangle the spurious and non-spurious associations,
yielding classifiers that hold up better
when spurious associations do not transport out of domain. 
We employ crowd workers not \emph{to label} documents,
but rather \emph{to edit} them, manipulating the text
to make a targeted (counterfactual) class applicable.
For sentiment analysis, we direct the worker to
\emph{revise this negative movie review to make it positive, 
without making any gratuitous changes}. 
We might regard the second part of this directive 
as a least action principle,
ensuring that we perturb only those spans necessary 
to alter the applicability of the label.
For NLI, a $3$-class classification task
(\emph{entailment, contradiction, neutral}), 
we ask the workers to modify the premise 
while keeping the hypothesis intact, 
and vice versa, collecting edits 
corresponding to each of the (two) counterfactual classes. 
Using this platform, we collect thousands 
of counterfactually-manipulated examples 
for both sentiment analysis and NLI, 
extending the IMDb \citep{maas2011learning} 
and SNLI \citep{bowman2015large} datasets, respectively. 
The result is two new datasets (each an extension of a standard resource) 
that enable us to both probe fundamental properties of language 
and train classifiers less reliant on spurious signal.

We show that classifiers trained on original IMDb reviews 
fail on counterfactually-revised data and vice versa. 
We further show that spurious correlations in these datasets 
are even picked up by linear models.
However, augmenting the revised examples breaks up these correlations 
(e.g., genre ceases to be predictive of sentiment). 
For a Bidirectional LSTM \citep{graves2005framewise} 
trained on IMDb reviews, 
classification accuracy goes down from $79.3\%$ to $55.7\%$ 
when evaluated on original vs revised reviews. 
The same classifier trained on revised reviews 
achieves an accuracy of $89.1\%$ on revised reviews 
compared to $62.5\%$ on their original counterparts. 
These numbers go to $81.7\%$ and $92.0\%$ on original
and revised data, respectively, 
when the classifier is retrained on the combined dataset. 
Similar patterns are observed for linear classifiers. 
We discovered that BERT \citep{devlin2019bert} is more resilient 
to such drops in performance on sentiment analysis. 

Additionally, SNLI models appear to rely on spurious associations
as identified by \citet{gururangan2018annotation}. 
Our experiments show that when fine-tuned
on original SNLI sentence pairs,
BERT fails on pairs with revised premise and vice versa, 
suffering more than a $30$ point drop in accuracy. 
Fine-tuned on the combined set, 
BERT's performance improves significantly across all datasets. 
Similarly, a Bi-LSTM trained on (original) hypotheses alone 
can accurately classify $69\%$ of pairs correctly 
but performs worse than the blind classifier
when evaluated on the revised dataset. 
When trained on hypotheses only from the combined dataset, 
its performance is not appreciably better than random guessing.
\section{Related Work}
Several papers demonstrate cases where NLP systems appear not to learn 
what humans consider to be \emph{the difference that makes the difference}.
For example, otherwise state-of-the-art models have been shown to be vulnerable 
to synthetic transformations such as distractor phrases \citep{jia2017adversarial,wallace2019universal}, 
to misclassify paraphrased task \citep{iyyer2018adversarial,pfeiffer2019deep} 
and to fail on template-based modifications \citep{ribeiro2018semantically}. 
\citet{glockner_acl18} demonstrate that simply replacing words by synonyms or hypernyms, 
which should not alter the applicable label, nevertheless breaks ML-based NLI systems.
\citet{gururangan2018annotation} and \citet{poliak2018hypothesis} 
show that classifiers correctly classified 
the hypotheses alone in about $69\%$ of SNLI corpus. 
They further discover that 
crowd workers adopted specific annotation strategies 
and heuristics for data generation. 
\citet{chen2016thorough} identify similar issues exist with automatically-constructed benchmarks
for question-answering \citep{hermann2015teaching}.
\citet{kaushik2018much} discover 
that reported numbers in question-answering benchmarks 
could often be achieved by the same models when restricted to be blind 
either to the question or to the passages. 
\citet{dixon2018measuring,zhao2018gender} and \citet{kiritchenko2018examining} 
showed how imbalances in training data lead to unintended bias in the resulting models, 
and, consequently,
potentially unfair applications.
\citet{shendarling} substitute words to test the behavior of sentiment analysis algorithms 
in the presence of stylistic variation, finding that similar word pairs 
produce significant differences in sentiment score.

Several papers explore richer feedback mechanisms for classification.
Some ask annotators to highlight \emph{rationales},
spans of text indicative of the label \citep{zaidan2007using, zaidan2008modeling, poulis2017learning}.
For each document, \citeauthor{zaidan2007using} remove the \emph{rationales} 
to generate \emph{contrast} documents, learning classifiers 
to distinguish original documents from their \emph{contrasting} counterparts. 
While this feedback is easier to collect than ours,
how to leverage it for training deep NLP models,
where features are not neatly separated, remains less clear.

\citet{lu2018gender} programmatically alter text to invert gender bias
and combined the original and manipulated data 
yielding gender-balanced dataset for learning word embeddings. 
In the simplest experiments, they swap each gendered word
for its other-gendered counterpart.
For example, \emph{the doctor ran because he is late} 
becomes \emph{the doctor ran because she is late}.
However, they do not substitute names even if they co-refer to a gendered pronoun.
Building on their work, \citet{zmigrod-etal-2019-counterfactual} 
describe a data augmentation approach for 
mitigating gender stereotypes associated with 
animate nouns for morphologically-rich languages like Spanish and Hebrew.
They use a Markov random field 
to infer how the sentence must be modified 
while altering the grammatical gender of particular nouns 
to preserve morpho-syntactic agreement. 
In contrast, \citet{maudslay2019s} describe 
a method for probabilistic automatic in-place substitution 
of gendered words in a corpus. 
Unlike \citeauthor{lu2018gender}, 
they propose an explicit treatment of first names by pre-defining name-pairs for swapping, 
thus expanding \citeauthor{lu2018gender}'s list of gendered word pairs significantly. 

\section{Data Collection}
\begin{figure}[t]
\centering
    \includegraphics[width=\textwidth]{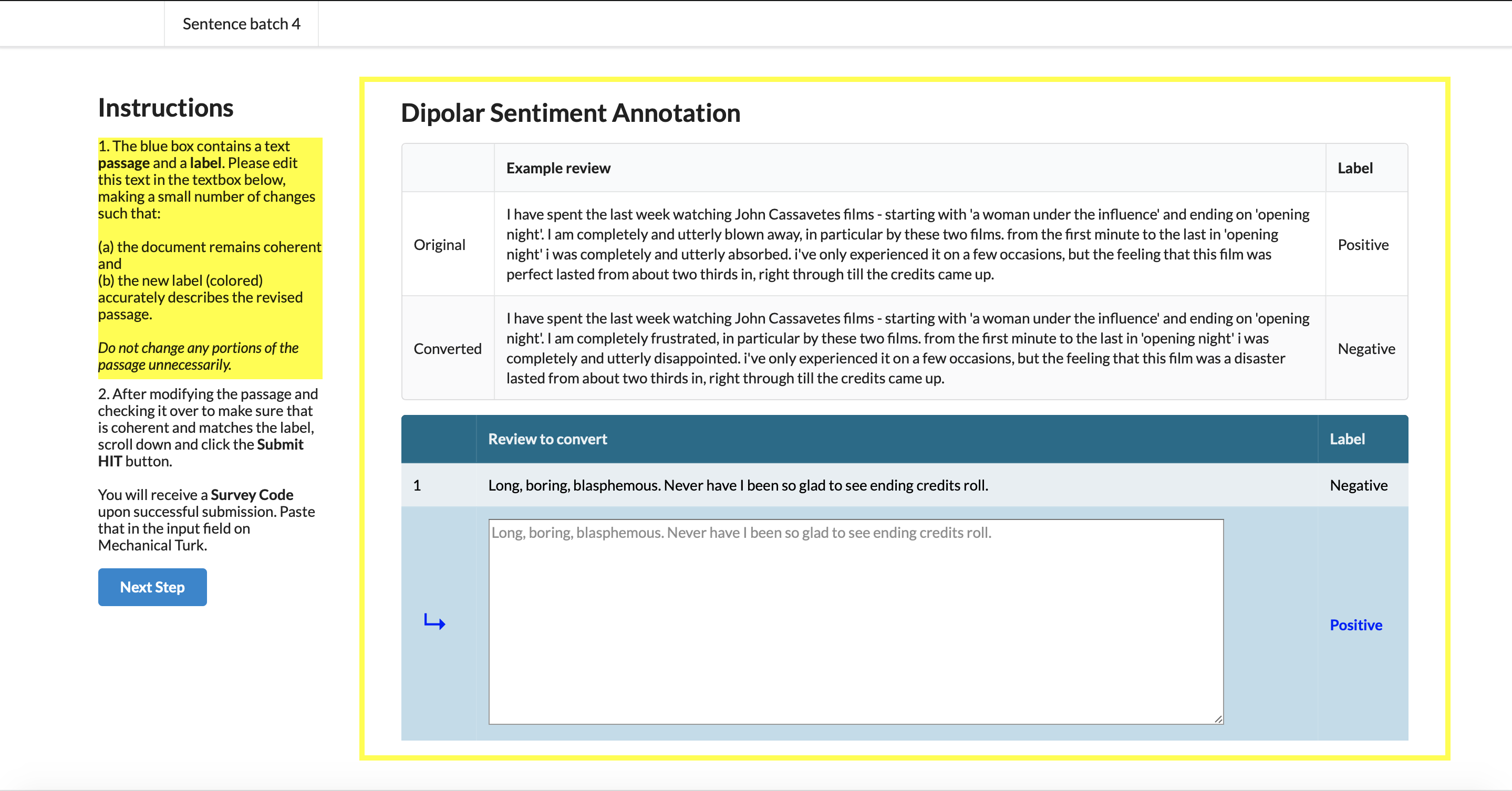}
\caption{
Annotation platform for collecting counterfactually annotated data for sentiment analysis
\label{fig:annotation_platform}}
\end{figure}
We use Amazon's Mechanical Turk crowdsourcing platform
to recruit editors to revise each document.
To ensure high quality of the collected data, 
we restricted the pool to U.S. residents that had already completed 
at least $500$ HITs and had an over $97\%$ HIT approval rate.
For each HIT, we conducted pilot tests 
to identify appropriate compensation per assignment, 
receive feedback from workers and revise our instructions accordingly. 
A total of $713$ workers contributed throughout the whole process, 
of which $518$ contributed edits reflected in the final datasets.

\textbf{Sentiment Analysis\quad}
The original IMDb dataset consists of $50k$ reviews 
divided equally across train and test splits. 
To keep the task of editing from growing unwieldy, 
we filter out the longest 20\% of reviews,
leaving $20k$ reviews in the train split
from which we randomly sample $2.5k$ reviews,
enforcing a $50$:$50$ class balance.
Following revision by the crowd workers, 
we partition this dataset 
into train/validation/test splits 
containing $1707$, $245$ and $488$ examples, respectively.
We present each review to two workers,
instructing them to revise the review such that 
(a) the counterfactual label applies;
(b) the document remains coherent;
and (c) no unecessary modifications are made.

Over a four week period, we manually inspected each generated review 
and rejected the ones that were outright wrong 
(sentiment was still the same or the review was a spam). 
After review, we rejected roughly $2\%$ of revised reviews. 
For $60$ original reviews, we did not approve any among 
the counterfactually-revised counterparts supplied by the workers.
To construct the new dataset, we chose one revised review (at random) 
corresponding to each original review.
In qualitative analysis, we identified eight common
patterns among the edits (Table~\ref{tab:edit_categories_sentiment}). 

By comparing original reviews to their counterfactually-revised counterparts
we gain insight into which aspects are causally relevant.
To analyze inter-editor agreement, we mark indices
corresponding to replacements and insertions,
representing the edits in each original review by a binary vector. 
Using these representations, we compute the Jaccard similarity 
between the two reviews (Table~\ref{tab:percent_inter_annotator}),
finding it to be negatively correlated with the length of the review.

\begin{table}[t!]
 \centering
  \begin{center} 
  \renewcommand{\arraystretch}{1.2}
  \caption{Percentage of inter-editor agreement for counterfactually-revised movie reviews\label{tab:percent_inter_annotator}}
  \setlength{\tabcolsep}{4pt}
  \begin{tabular}{ l c c c c c c c c}
    \toprule
    & \multicolumn{7}{c}{Number of tokens}\\
    Type & 0-50 & 51-100 & 101-150 & 151-200 & 201-250 & 251-300 & 301-329 & Full \\
    \midrule
    Replacement & $35.6$ & $25.7$ & $20.0$ & $17.2$  & $15.0$ & $14.8$ & $11.6$ & $19.3$ \\
    Insertion & $27.7$ & $20.8$ & $14.4$ & $12.2$& $11.0$ & $11.5$ & $07.6$ & $14.3$ \\
    Combined & $41.6$ & $32.7$ & $26.3$ & $23.4$& $21.6$ & $20.3$ & $16.2$ & $25.5$ \\
\bottomrule
  \end{tabular}
  \end{center}
  \vspace{-10px}
\end{table}

\begin{table}[t!]
\centering
 \renewcommand{\arraystretch}{1.2}
  \caption{Most prominent categories of edits performed by humans for sentiment analysis (Original/Revised, in order). Red spans were replaced by Blue spans.
  \label{tab:edit_categories_sentiment}}
  \begin{tabularx}{13.9cm}{lX}
    \toprule
    Types of Revisions & Examples \\
    \midrule
    Recasting \emph{fact} as \emph{hoped for} & The world of Atlantis, hidden beneath the earth's core, is fantastic \newline The world of Atlantis, hidden beneath the earth's core is \textbf{\textcolor{blue}{supposed}} to be fantastic\\
    
    Suggesting sarcasm & thoroughly captivating \textbf{\textcolor{red}{thriller-drama, taking a deep and realistic}} view \newline thoroughly mind numbing \textbf{\textcolor{blue}{``thriller-drama'', taking a ``deep'' and ``realistic'' (who are they kidding?)}} view \\
    Inserting modifiers & The presentation of simply Atlantis' landscape and setting \newline The presentation of Atlantis' \textbf{\textcolor{blue}{predictable}} landscape and setting \\
    Replacing modifiers &  ``Election'' is a highly fascinating and thoroughly \textbf{\textcolor{red}{captivating}} thriller-drama \newline ``Election'' is a highly expected and thoroughly \textbf{\textcolor{blue}{mind numbing}} ``thriller-drama'' \\
    Inserting phrases & Although there's hardly any action, the ending is still shocking. \newline Although there's hardly any action \textbf{\textcolor{blue}{(or reason to continue watching past 10 minutes)}}, the ending is still shocking. \\
    Diminishing via qualifiers & which, while usually containing some reminder of harshness, become \textbf{\textcolor{red}{more and more intriguing}}. \newline which, usually containing some reminder of harshness, became \textbf{\textcolor{blue}{only slightly more intriguing}}. \\
    Differing perspectives & Granted, \textbf{\textcolor{red}{not all of the story makes full sense}}, but the film doesn't feature any amazing new computer-generated visual effects. \newline Granted, \textbf{\textcolor{blue}{some of the story makes sense}}, but the film doesn't feature any amazing new computer-generated visual effects. \\
    
    Changing ratings & one of the worst ever scenes in a sports movie. \textbf{\textcolor{red}{3 stars out of 10}}. \newline one of the wildest ever scenes in a sports movie. \textbf{\textcolor{blue}{8 stars out of 10}}.\\
\bottomrule
  \end{tabularx}
  \vspace{0px}
\end{table}

\textbf{Natural Language Inference\quad}
Unlike sentiment analysis, SNLI is $3$-way classification task, 
with inputs consisting of two sentences,
a \emph{premise} and a \emph{hypothesis} 
and the three possible labels being \emph{entailment}, 
\emph{contradiction}, and \emph{neutral}.
The label is meant to describe the relationship 
between the facts stated in each sentence.
We randomly sampled $1750$, $250$, and $500$ pairs
from the train, validation, and test sets of SNLI respectively, 
constraining the new data to have balanced classes.
In one HIT, we asked workers to revise the hypothesis 
while keeping the premise intact, 
seeking edits corresponding to each of the two counterfactual classes. 
We refer to this data as Revised Hypothesis (RH). 
In another HIT, we asked workers to revise the original premise,
while leaving the original hypothesis intact, 
seeking similar edits, 
calling it Revised Premise (RP).

Following 
data collection,
we employed a different set of workers 
to verify whether the given label accurately described 
the relationship between each premise-hypothesis pair. 
We presented each pair to three workers and performed a majority vote. 
When all three reviewers were in agreement, 
we approved or rejected the pair based on their decision, 
else, we verified the data ourselves. 
Finally, we only kept premise-hypothesis pairs 
for which we had valid revised data in both RP and RH, corresponding to both counterfactual labels. 
As a result, we
discarded $\approx 9\%$ data. 
RP and RH, each comprised of $3332$ pairs in train, 
$400$ in validation, and $800$ in test, leading to a total of 
$6664$ pairs in train, $800$ in validation, and $1600$ in test in the revised dataset. In qualitative analysis, 
we identified some common
patterns among hypothesis and premise edits (Table~\ref{tab:edit_categories_snli_hypothesis},~\ref{tab:edit_categories_snli_premise}).

We collected all data after IRB approval 
and measured the time taken to complete each HIT 
to ensure that all workers were paid
more than the federal minimum wage.
During our pilot studies, 
workers spent roughly $5$ minutes per revised review, and
$4$ minutes per revised sentence (for NLI). 
We paid workers $\$0.65$ per revision, 
and $\$0.15$ per verification,
totalling $\$10778.14$ for the study.

\begin{table}[t!]
\centering
 \renewcommand{\arraystretch}{1.2}
  \caption{Analysis of edits performed by humans for NLI hypotheses. P denotes \emph{Premise}, OH denotes \emph{Original Hypothesis}, and NH denotes \emph{New Hypothesis}.
  \label{tab:edit_categories_snli_hypothesis}}
  \begin{tabularx}{13.9cm}{lX}
    \toprule
    Types of Revisions & Examples \\
    \midrule
    Modifying/removing actions & \textbf{P:} A young dark-haired woman crouches on the banks of a river while washing dishes.\newline
    \textbf{OH:} A woman washes dishes in the river \textbf{\textcolor{red}{while camping}}. (Neutral)\newline
    \textbf{NH:} A woman washes dishes in the river. (Entailment)\\
    
    Substituting entities & \textbf{P:} Students are inside of a lecture hall.\newline
    \textbf{OH:} Students are \textbf{\textcolor{red}{indoors}}. (Entailment)\newline
    \textbf{NH:} Students are \textbf{\textcolor{blue}{on the soccer field}}. (Contradiction)
 \\
    Adding details to entities & \textbf{P:} An older man with glasses raises his eyebrows in surprise.\newline
    \textbf{OH:} The man \textbf{\textcolor{red}{has no glasses}}. (Contradiction)\newline
    \textbf{NH:} The man \textbf{\textcolor{blue}{wears bifocals}}. (Neutral)
 \\
    Inserting relationships & \textbf{P:} A blond woman speaking to a brunette woman with her arms crossed.\newline
    \textbf{OH:} A woman is talking to \textbf{\textcolor{red}{another woman}}. (Entailment)\newline
    \textbf{NH:} A woman is talking to \textbf{\textcolor{blue}{a family member}}. (Neutral)
 \\
    Numerical modifications & \textbf{P:} Several farmers bent over working on the fields while lady with a baby and four other children accompany them.\newline
    \textbf{OH:} The lady has \textbf{\textcolor{red}{three}} children. (Contradiction)\newline
    \textbf{NH:} The lady has \textbf{\textcolor{blue}{many}} children. (Entailment)
 \\
    Using/Removing negation & \textbf{P:} An older man with glasses raises his eyebrows in surprise.\newline
    \textbf{OH:} The man \textbf{\textcolor{red}{has no}} glasses. (Contradiction)\newline
    \textbf{NH:} The man \textbf{\textcolor{blue}{wears}} glasses. (Entailment)
 \\
    Unrelated hypothesis & \textbf{P:} A female athlete in crimson top and dark blue shorts is running on the street.\newline
    \textbf{OH:} A woman is \textbf{\textcolor{red}{sitting on}} a white couch. (Contradiction)\newline
    \textbf{NH:} A woman \textbf{\textcolor{blue}{owns}} a white couch. (Neutral)
\\
\bottomrule
  \end{tabularx}
\end{table}

\begin{table}[t!]
\centering
 \renewcommand{\arraystretch}{1.2}
  \caption{Analysis of edits performed by humans for NLI premises. OP denotes \emph{Original Premise}, NP denotes \emph{New Premise}, and H denotes \emph{Hypothesis}.
  \label{tab:edit_categories_snli_premise}}
  \begin{tabularx}{13.9cm}{lX}
    \toprule
    Types of Revisions & Examples \\
    \midrule
    Introducing direct evidence & \textbf{OP:} Man walking with tall buildings with reflections behind him. (Neutral)\newline
    \textbf{NP:} Man walking \textbf{\textcolor{blue}{away from his friend}}, with tall buildings with reflections behind him. (Contradiction)\newline
    \textbf{H:} The man was walking to meet a friend.
\\
    Introducing indirect evidence & \textbf{OP:} An Indian man standing on the bank of a river. (Neutral)\newline
    \textbf{NP:} An Indian man standing \textbf{\textcolor{blue}{with only a camera}} on the bank of a river. (Contradiction)\newline
    \textbf{H:} He is fishing.
\\
    Substituting entities & \textbf{OP:} A young man in front of a \textbf{\textcolor{red}{grill}} laughs while pointing at something to his left. (Entailment) \newline
    \textbf{NP:} A young man in front of a \textbf{\textcolor{blue}{chair}} laughs while pointing at something to his left. (Neutral)\newline
    \textbf{H:} A man is outside\\
    Numerical modifications & \textbf{OP:} The exhaustion in the woman's face while she continues to ride her bicycle in the competition. (Neutral)\newline
    \textbf{NP:} The exhaustion in the woman's face while she continues to ride her bicycle in the competition \textbf{\textcolor{blue}{for people above 7 ft}}. (Entailment)\newline
    \textbf{H:} A tall person on a bike
 \\

    Reducing evidence & \textbf{OP:} The girl in yellow shorts and white jacket has a tennis ball \textbf{\textcolor{red}{in her left pocket}}. (Entailment)\newline
    \textbf{NP:} The girl in yellow shorts and white jacket has a tennis ball. (Neutral)\newline
    \textbf{H:} A girl with a tennis ball in her pocket.
 \\
    Using abstractions & \textbf{OP:} An elderly \textbf{\textcolor{red}{woman}} in a crowd pushing a wheelchair. (Entailment)\newline
    \textbf{NP:} An elderly \textbf{\textcolor{blue}{person}} in a crowd pushing a wheelchair. (Neutral)\newline
    \textbf{H:} There is an elderly woman in a crowd.
 \\
    Substituting evidence & \textbf{OP:} A woman is \textbf{\textcolor{red}{cutting something with scissors}}. (Entailment)\newline
    \textbf{NP:} A woman is \textbf{\textcolor{blue}{reading something about scissors}}. (Contradiction)\newline
    \textbf{H:} A woman uses a tool\\

\bottomrule
  \end{tabularx}
  \vspace{0px}
\end{table}
\section{Models}
Our experiments rely on the following five models: Support Vector Machines (SVMs), Na\"ive Bayes (NB) classifiers, Bidirectional Long Short-Term Memory Networks \citep[Bi-LSTMs;][]{graves2005framewise}, ELMo models with LSTM, and fine-tuned BERT models \citep{devlin2019bert}.
For brevity, we discuss only implementation details necessary for reproducibility.

\textbf{Standard Methods\quad} 
We use \texttt{scikit-learn} \citep{scikit-learn} 
implementations of SVMs and Na\"ive Bayes
for sentiment analysis. 
We train these models on TF-IDF bag of words feature representations of the reviews. 
We identify parameters for 
both
classifiers 
using grid search conducted over the validation set. 

\textbf{Bi-LSTM\quad} 
When training Bi-LSTMs for sentiment analysis, 
we restrict the vocabulary to the most frequent $20k$ tokens, replacing out-of-vocabulary tokens by \texttt{UNK}. 
We fix the maximum input length at $300$ tokens and pad smaller reviews.
Each token is represented by a randomly-initialized $50$-dimensional embedding.
Our model consists of a bidirectional LSTM (hidden dimension $50$) 
with recurrent dropout 
(probability $0.5$) and global max-pooling 
following the embedding layer. 
To generate output, 
we feed this (fixed-length) representation through
a fully-connected hidden layer with ReLU \citep{nair2010rectified} activation 
(hidden dimension $50$), 
and then a fully-connected output layer 
with softmax activation. 
We train all models for a maximum of $20$ epochs 
using Adam \citep{kingma2014adam}, 
with a learning rate of $\expnumber{1}{-3}$ and a batch size of $32$.  
We apply early stopping when validation loss does not decrease for $5$ epochs.
We also experimented with a larger Bi-LSTM which led to overfitting.
We use the architecture due to \citet{poliak2018hypothesis} to evaluate hypothesis-only baselines.\footnote{\href{https://github.com/azpoliak/hypothesis-only-NLI}{https://github.com/azpoliak/hypothesis-only-NLI}}

\textbf{ELMo-LSTM\quad}
We compute contextualized word representations (ELMo) using character-based word representations and bidirectional LSTMs \citep{peters2018deep}. The module outputs a $1024$-dimensional weighted sum of representations from the $3$ Bi-LSTM layers used in ELMo.
We represent each word by a $128$-dimensional embedding concatenated to the resulting $1024$-dimensional ELMo representation, leading to a $1152$-dimensional hidden representation. 
Following Batch Normalization, this is passed through an LSTM (hidden size $128$) with recurrent dropout (probability $0.2$). The output from this LSTM is then passed to a fully-connected output layer with softmax activation. We train this model for up to $20$ epochs with same early stopping criteria as for Bi-LSTM, using the Adam optimizer with a learning rate of $\expnumber{1}{-3}$ and a batch size of $32$.

\textbf{BERT\quad} 
We use an off-the-shelf uncased BERT Base 
model, fine-tuning for each task.\footnote{\href{https://github.com/huggingface/pytorch-transformers}{https://github.com/huggingface/pytorch-transformers}}
To account for BERT's sub-word tokenization,
we set the maximum token length is set at 
$350$ for sentiment analysis and $50$ for NLI.
We fine-tune BERT up to $20$ epochs 
with same early stopping criteria as for Bi-LSTM,
using the BERT Adam optimizer with a batch size of $16$ 
(to fit on a Tesla V-$100$ GPU).
We found learning rates of $\expnumber{5}{-5}$ and $\expnumber{1}{-5}$ 
to work best for sentiment analysis and NLI respectively.
\section{Experimental Results}
\paragraph{Sentiment Analysis}
We find that for sentiment analysis, 
linear models trained on the original $1.7k$ reviews 
achieve $80\%$ accuracy when evaluated on original reviews 
but only $51\%$ (level of random guessing) on revised reviews
(Table \ref{tab:sentiment_results}).
Linear models trained on revised reviews 
achieve $91\%$ accuracy on revised reviews 
but only $58.3\%$ on the original test set.
We see similar pattern for Bi-LSTMs 
where accuracy drops substantially in both directions.
Interestingly, while BERT models suffer drops too,
they are less pronounced, perhaps a benefit 
of the exposure to a larger dataset 
where the spurious patterns may not have held.
Classifiers trained on combined datasets perform well on both, 
often within $\approx 3$ pts of models 
trained on the same amount of data
taken only from the original distribution. 
Thus, there may be a price to pay 
for breaking the reliance on spurious associations,
but it may not be substantial.

We also conduct experiments to evaluate our sentiment models
vis-a-vis their generalization out-of-domain to new domains. 
We evaluate models on Amazon reviews \citep{ni-etal-2019-justifying} 
on data aggregated over six genres: \emph{beauty, fashion, appliances, giftcards, magazines,} and \emph{software}, the Twitter sentiment dataset \citep{rosenthal-etal-2017-semeval},\footnote{We use the development set as test data is not public.} 
and Yelp reviews released as part of the Yelp dataset challenge.
We show that in almost all cases, 
models trained on the counterfactually-augmented IMDb dataset 
perform better than models trained on comparable quantities of original data.

To gain intuition about what is learnable absent the edited spans,
we tried training several models on passages
where the edited spans have been removed
from training set sentences (but not test set).
SVM, Na\"ive Bayes, and Bi-LSTM achieve $57.8\%, 59.1\%, 60.2\%$ accuracy, respectively, 
on this task.
Notably, these passages are predictive of the (true) label
despite being semantially compatible with the counterfactual label. 
However, BERT performs worse than random guessing. 

In one simple demonstration of the benefits of our approach,
we note that seemingly irrelevant words such 
as: \textcolor{blue}{\textit{romantic, will, my, has, especially, life, works, both, it, its, lives}} and \textcolor{blue}{\textit{gives}} (correlated with positive sentiment), and \textcolor{red}{\textit{horror, own, jesus, cannot, even, instead, minutes, your, effort, script, seems}} and \textcolor{red}{\textit{something}} (correlated with negative sentiment)
are picked up as high-weight features by linear models 
trained on either original or revised reviews as top predictors.
However, because humans never edit these during revision
owing to their lack of semantic relevance,
combining the original and revised datasets breaks these associations 
and these terms cease to be predictive of sentiment (Fig \ref{fig:svm_features_30}). 
Models trained on original data but at the same scale as combined data
are able to perform slightly better on the original test set
but still fail on the revised reviews.
All models trained on $19k$ original reviews 
receive a slight boost in accuracy on revised data 
(except Na\"ive Bayes), 
yet their performance significantly worse compared to specialized models. 
Retraining models on a combination of the original $19k$ reviews
with revised $1.7k$ reviews leads to significant increases in accuracy
for all models on classifying revised reviews, 
while slightly improving the accuracy on classifying the original reviews. 
This underscores the importance of including 
counterfactually-revised examples in training data.

\begin{figure}[t!]
    \begin{subfigure}[b]{.32\textwidth}
         \centering
         \includegraphics[width=\linewidth]{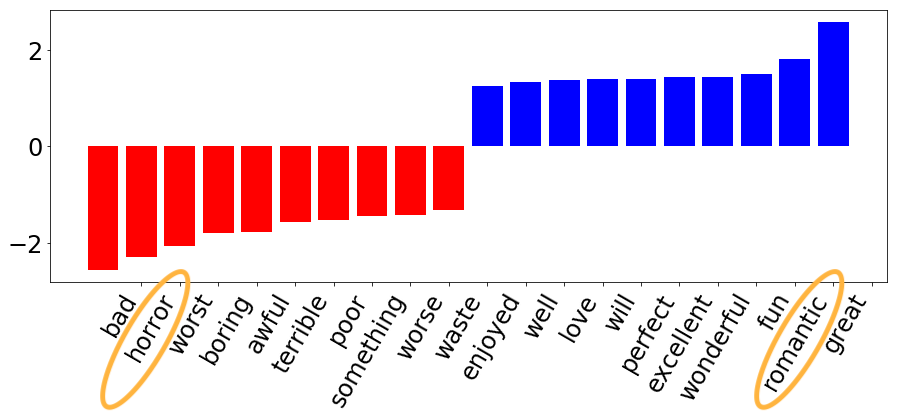}
         \caption{Trained on the original dataset}
         \label{fig:svm_orig}
     \end{subfigure}
      \begin{subfigure}[b]{.32\textwidth}
          \centering
          \includegraphics[width=\linewidth]{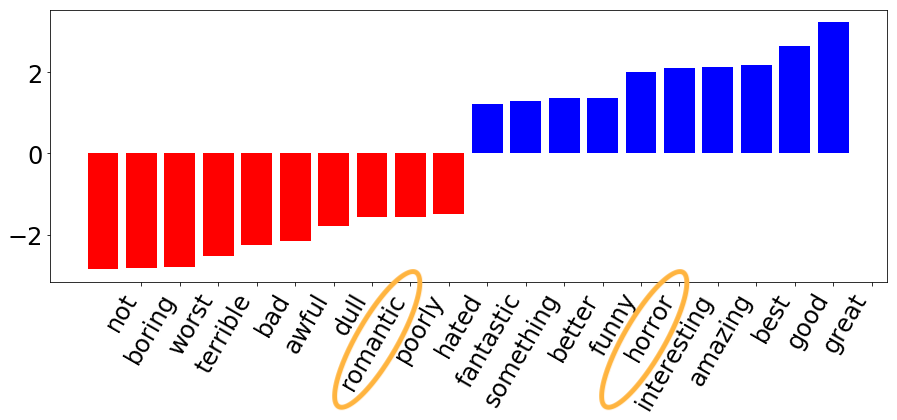}
          \caption{Trained on the revised dataset}
          \label{fig:svm_new}
      \end{subfigure}
     \begin{subfigure}[b]{.33\textwidth}
         \centering
         \includegraphics[width=\textwidth]{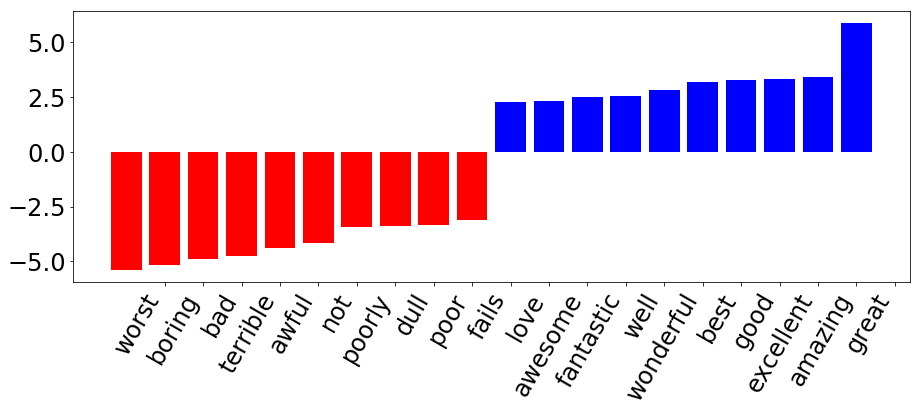}
         \caption{Trained on combined dataset}
         \label{fig:svm_augmented}
     \end{subfigure}
\caption{
Most important features learned by an SVM classifier trained on TF-IDF bag of words.
\label{fig:svm_features}}
\end{figure}

\begin{table}[t!]
  \begin{center}
  \renewcommand{\arraystretch}{1.2}
  \caption{Accuracy of various models for sentiment analysis trained with various datasets. Orig. denotes \emph{original}, \emph{Rev.} denotes revised, and \emph{Orig. - Edited} denotes the original dataset where the edited spans have been removed. 
  \label{tab:sentiment_results}}
  \begin{tabular}{ l c c c c c c c c c c}
    \toprule
    Training data & \multicolumn{2}{c}{SVM} & \multicolumn{2}{c}{NB} & \multicolumn{2}{c}{ELMo} & \multicolumn{2}{c}{Bi-LSTM} & \multicolumn{2}{c}{BERT} \\
    \midrule
    & O & R & O & R & O & R & O & R & O & R  \\
    \midrule
    Orig. ($1.7k$) & $\textbf{80.0}$& $51.0$&$\textbf{74.9}$ &$47.3$& $\textbf{81.9}$ & $66.7$ & $\textbf{79.3}$ & $55.7$ & $\textbf{87.4}$ & $82.2$\\
    Rev. ($1.7k$) & $58.3$ & $\textbf{91.2}$ & $50.9$ & $\textbf{88.7}$ & $63.8$ & $\textbf{82.0}$ & $62.5$ & $\textbf{89.1}$ & $80.4$ & $\textbf{90.8}$ \\
    Orig. $-$ Edited & $57.8$ & $-$ & $59.1$& $-$ & $50.3$& $-$ & $60.2$ & $-$ & $49.2$ & $-$\\
    \midrule
    Orig. \& Rev. ($3.4k$) & $83.7$ & $\textbf{87.3}$ & $\textbf{86.1}$& $\textbf{91.2}$ & $\textbf{85.0}$ & $\textbf{92.0}$& $\textbf{81.5}$ & $\textbf{92.0}$ & $88.5$ & $\textbf{95.1}$ \\
    Orig. ($3.4k$) & $\textbf{85.1}$ & $54.3$ & $82.4$& $48.2$ & $82.4$ & $61.1$ & $80.4$ & $59.6$ & $\textbf{90.2}$ & $86.1$\\
    \midrule
    Orig. ($19k$) & $\textbf{87.8}$ & $60.9$ & $84.3$ & $42.8$ & $86.5$ & $64.3$ & $86.3$ & $68.0$ & $93.2$ & $88.3$\\
    Orig. ($19k$) \& Rev. & $\textbf{87.8}$ & $\textbf{76.2}$ & $\textbf{85.2}$ & $\textbf{48.4}$ & $\textbf{88.3}$ & $\textbf{84.6}$ & $\textbf{88.7}$ & $\textbf{79.5}$ & $\textbf{93.2}$ & $\textbf{93.9}$\\
\bottomrule
  \end{tabular}
  \end{center}
\end{table}

\begin{table}[t!]
 \begin{center}
  \renewcommand{\arraystretch}{1.2}
  \caption{Accuracy of various sentiment analysis models on out-of-domain data}\label{tab:sentiment_out}
  \begin{tabular}{ l c c c c c}
    \toprule
    Training data & SVM & NB & ELMo & Bi-LSTM & BERT \\
    \midrule
    \multicolumn{6}{c}{Accuracy on Amazon Reviews}\\
    \midrule
    Orig. \& Rev. ($3.4k$) & $\textbf{77.1}$ & $\textbf{82.6}$ & $\textbf{78.4}$ & $\textbf{82.7}$ & $\textbf{85.1}$ \\
    Orig. ($3.4k$) & $74.7$ & $66.9$ & $\textbf{79.1}$ & $65.9$ & $80.0$ \\
    \midrule
    \multicolumn{6}{c}{Accuracy on Semeval 2017 (Twitter)}\\
    \midrule
    Orig. \& Rev. ($3.4k$) & $\textbf{66.5}$ & $\textbf{73.9}$ & $\textbf{70.0}$ & $\textbf{68.7}$ & $\textbf{82.9}$ \\
    Orig. ($3.4k$) & $61.2$ & $64.6$ & \textbf{69.5} & $55.3$ & $79.3$ \\
    \midrule
    \multicolumn{6}{c}{Accuracy on Yelp Reviews}\\
    \midrule
    Orig. \& Rev. ($3.4k$) & $\textbf{87.6}$ & $\textbf{89.6}$ & $\textbf{87.2}$ & $\textbf{86.2}$ & $\textbf{89.4}$ \\
    Orig. ($3.4k$) & $81.8$ & $77.5$ & $82.0$ & $78.0$ & $85.3$ \\
\bottomrule
  \end{tabular}
  \end{center}
\end{table}

\begin{table}[t!]
 \begin{center}
  \renewcommand{\arraystretch}{1.2}
  \caption{Accuracy of BERT on NLI with various train and eval sets.}\label{tab:bert_nli}
  \begin{tabular}{ l c c c c}
    \toprule
    Train/Eval & Original & RP & RH & RP \& RH \\
    \midrule
    Original ($1.67k$) & $72.2$ & $39.7$ & $59.5$ & $49.6$ \\
    \midrule
    Revised Premise (RP; $3.3k$) & $50.6$ & $66.3$ & $50.1$ & $58.2$ \\
    Revised Hypothesis (RH; $3.3k$)& $71.9$ & $47.4$ & $67.0$ & $57.2$ \\
    \midrule
    RP \& RH ($6.6k$) & $64.7$ & $64.6$  & $67.8$ & $66.2$ \\
    Original w/ RP \& RH ($8.3k$) & $73.5$ & $\textbf{64.6}$ & $\textbf{69.6}$ & $\textbf{67.1}$ \\
    Original ($8.3k$) & $\textbf{77.8}$ & $44.6$ & $66.1$ & $55.4$ \\
    \midrule
    Original ($500k$) & $90.4$ & $54.3$ & $74.3$ & $64.3$ \\
\bottomrule
  \end{tabular}
  \end{center}
\end{table}

\begin{table}[t!]
 \begin{center}
  \renewcommand{\arraystretch}{1.2}
  \caption{Accuracy of Bi-LSTM classifier trained on hypotheses only\label{tab:hypoth_only}}
  \begin{tabular}{ l c c c c}
    \midrule
    Train/Test & Original & RP & RH & RP \& RH \\
    \midrule
    Majority class & $34.7$ & $34.6$ & $34.6$ & $34.6$\\
    \midrule
    RP \& RH ($6.6k$)  & $\textbf{32.4}$ & $\textbf{35.1}$  & $\textbf{33.4}$ & $\textbf{34.2}$ \\
    Original w/ RP \& RH ($8.3k$) & $44.0$ & $25.8$ & $43.2$ & $\textbf{34.5}$ \\
    Original ($8.3k$) & $60.2$ & $20.5$ & $46.6$ & $\textbf{33.6}$ \\
    Original ($500k$) & $69.0$ & $15.4$ & $53.2$ & $\textbf{34.3}$ \\
\bottomrule
  \end{tabular}
  \end{center}
\end{table}

\begin{table}[t!]
 \begin{center}
  \renewcommand{\arraystretch}{1.2}
  \caption{Accuracy of models trained to differentiate between original and revised data\label{tab:discriminator}}
  \begin{tabular}{ l c c c}
    \toprule
    Model & IMDb & SNLI/RP & SNLI/RH \\
    \midrule
    Majority class & $50.0$ & $66.7$  & $66.7$ \\
    \midrule
    SVM & $67.4$ & $46.6$ & $51.0$ \\
    NB & $69.2$ & $\textbf{66.7}$ & $66.6$ \\
    BERT & $\textbf{77.3}$ & $64.8$ & $\textbf{69.7}$ \\
\bottomrule
  \end{tabular}
  \end{center}
  \vspace{-10px}
\end{table}

\textbf{Natural Language Inference\quad} 
Fine-tuned on $1.67k$ original sentence pairs, 
BERT achieves $72.2\%$ accuracy on SNLI dataset 
but it is only able to accurately classify
$39.7\%$ sentence pairs from the RP set (Table~\ref{tab:bert_nli}). 
Fine-tuning BERT on the full SNLI training set 
($500k$ sentence pairs) results in similar behavior. 
Fine-tuning it on RP sentence pairs improves its accuracy to $66.3\%$ 
on RP but causes a drop of roughly $20$ pts on SNLI. 
On RH sentence pairs, this results in an accuracy 
of $67\%$ on RH and $71.9\%$ on SNLI test set but $47.4\%$ on the RP set. 
To put these numbers in context, 
each individual hypothesis sentence in RP 
is associated with two labels, 
each in the presence of a different premise. 
A model that relies on hypotheses only would at best perform slightly better
than choosing the majority class when evaluated on this dataset.
However, fine-tuning BERT on a combination of RP and RH 
leads to consistent performance on all datasets 
as the dataset design forces models to look at both premise and hypothesis. 
Combining original sentences with RP and RH 
improves these numbers even further. 
We compare this with the performance obtained 
by fine-tuning it on $8.3k$ sentence pairs sampled from SNLI training set, 
and show that while the two perform roughly within $4$ pts of each other 
when evaluated on SNLI, the former outperforms latter on both RP and RH. 

To further isolate this effect, Bi-LSTM trained on SNLI hypotheses only 
achieves $69\%$ accuracy on SNLI test set, 
which drops to $44\%$ if it is retrained 
on combination of original, RP and RH data (Table~\ref{tab:hypoth_only}).
Note that this combined dataset consists of five variants of each original premise-hypothesis pair. 
Of these five pairs, three consist of the same hypothesis sentence,
each associated with different truth value given the respective premise.
Using these hypotheses only would provide conflicting feedback
to a classifier during training,
thus causing the drop in performance.
Further, we notice that the gain of the latter over majority class baseline 
comes primarily from the original data, 
as the same model retrained only on RP and RH data experiences a further drop of $11.6\%$ in accuracy, 
performing worse than just choosing the majority class at all times.

One reasonable concern might be that our models 
would simply distinguish whether an example 
were from the original or revised dataset
and thereafter treat them differently.
The fear might be that our models would exhibit a hypersensitivity
(rather than insensitivity) to domain.
To test the potential for this behavior, 
we train several models to distinguish between
original and revised data (Table~\ref{tab:discriminator}). 
BERT identifies original reviews from revised reviews with $77.3\%$ accuracy.
In case of NLI, BERT and Na\"ive Bayes perform roughly within $3$ pts 
of the majority class baseline ($66.7\%$) whereas SVM performs substantially worse.
\section{Conclusion}
By leveraging humans not only to provide labels 
but also to intervene upon the data, revising documents 
to accord with various labels, 
we can elucidate the difference that makes a difference. 
Moreover, we can leverage the augmented data to train classifiers 
less dependent on spurious associations. 
Our study demonstrates the promise of leveraging human-in-the-loop feedback
to disentangle the spurious and non-spurious associations,
yielding classifiers that hold up better
when spurious associations do not transport out of domain. 
Our methods appear useful on both sentiment analysis and NLI,
two contrasting tasks. 
In sentiment analysis, expressions of opinion matter more than stated facts,
while in NLI this is reversed.
SNLI poses another challenge in that it is a $3$-class classification task using two input sentences. 
In future work, we will extend these techniques, 
leveraging humans in the loop to build more 
robust systems for question answering and summarization. 
\section*{Acknowledgements}
The authors are grateful to Amazon AWS and NVIDIA
for providing GPUs to conduct the experiments, 
Salesforce Research and Facebook AI 
for their generous grants 
that made the data collection possible, 
Sina Fazelpour, Sivaraman Balakrishnan, 
Shruti Rijhwani, Shruti Palaskar, Aishwarya Kamath, 
Michael Collins, Rajesh Ranganath and Sanjoy Dasgupta 
for their valuable feedback, and Tzu-Hsiang Lin 
for his generous help in creating the data collection platform.
We also thank Abridge AI, UPMC, the Center for Machine Learning in Health,
and the AI Ethics and Governance Fund
for their support of our broader research on robust machine learning.

\bibliography{refs}
\bibliographystyle{iclr2020_conference}

\newpage
\appendix
\section*{Appendix}
\begin{table}[h!]
  \begin{center}
  \renewcommand{\arraystretch}{1.2}
  \caption{Most frequent insertions/deletions by human annotators for sentiment analysis. 
  \label{tab:insert_delete_sent}}
  \begin{tabularx}{13.9cm}{l X X}
    \toprule
    Revision & Removed words & Inserted words \\
    \midrule
    Positive to Negative & \emph{movie, film, great, like, good, really, would, see, story, love} & \emph{movie, film, one, like, bad, would, really, even, story, see} \\
    Negative to Positive & \emph{bad, even, worst, waste, nothing, never, much, would, like, little} & \emph{great, good, best, even, well, amazing, much, many, watch, better} \\
    \bottomrule
  \end{tabularx}
  \end{center}
\end{table}

\begin{table}[t!]
  \begin{center}
  \renewcommand{\arraystretch}{1.2}
  \caption{Most frequent insertions/deletions by human annotators for SNLI. 
  \label{tab:insert_delete_nli}}
  \begin{tabularx}{13.9cm}{l X X}
    \toprule
    Revision & Removed words & Inserted words \\
    \midrule
    \multicolumn{3}{>{\centering\hsize=\dimexpr3\hsize+3\tabcolsep+\arrayrulewidth\relax}X}{Revising Premise}\\
    \midrule
    Entailment to Neutral & \emph{woman, walking, man, blue, sitting, men, girl, standing, looking, running} & \emph{person, near, child, something, together, people, tall, vehicle, wall, holding} \\
    Neutral to Entailment & \emph{man, street, black, water, little, front, young, playing, woman, two} & \emph{waiting, couple, playing, running, getting, making, tall, game, black, happily}\\
    Entailment to Contradiction & \emph{blue, people, standing, girl, front, street, red, young, sitting, band} & \emph{sitting, standing, inside, young, women, child, red, men, sits, one} \\
    Contradiction to Entailment & \emph{sitting, man, walking, black, blue, people, red, standing, white, street} & \emph{man, sitting, sleeping, woman, sits, eating, playing, park, two, standing} \\
    Neutral to Contradiction & \emph{man, woman, people, boy, black, red, standing, young, two, water} & \emph{man, woman, boy, men, alone, sitting, girl, dog, three, one} \\
    Contradiction to Neutral & \emph{man, sitting, black, blue, walking, red, standing, street, white, street} & \emph{man, sitting, woman, people, person, near, something, something, sits, black} \\
    \midrule
    \multicolumn{3}{>{\centering\hsize=\dimexpr3\hsize+3\tabcolsep+\arrayrulewidth\relax}X}{Revising Hypothesis}\\
    \midrule
    Entailment to Neutral & \emph{man, wearing, white, blue, black, shirt, one, young, people, woman} & \emph{people, there, playing, man, person, wearing, outside, two, old, near} \\
    Neutral to Entailment & \emph{white, wearing, shirt, black, blue, man, two, standing, young, red} & \emph{playing, wearing, man, two, there, woman, people, men, near, person}\\
    Entailment to Contradiction & \emph{man, wearing, white, blue, black, two, shirt, one, young, people} & \emph{people, man, woman, playing, no, inside, person, two, wearing, women} \\
    Contradiction to Entailment & \emph{wearing, blue, black, man, white, two, red, shirt, young, one} & \emph{people, there, man, two, wearing, playing, people, men, woman, outside} \\
    Neutral to Contradiction & \emph{white, man, wearing, shirt, black, blue, two, standing, woman, red} & \emph{woman, man, there, playing, two, wearing, one, men, girl, no} \\
    Contradiction to Neutral & \emph{wearing, blue, black, man, white, two, red, sitting, young, standing} & \emph{people, playing, man, woman, two, wearing, near, tall, men, old} \\
    \bottomrule
  \end{tabularx}
  \end{center}
\end{table}

\begin{figure}[t]
    \begin{subfigure}[b]{\textwidth}
         \centering
         \includegraphics[width=0.8\linewidth]{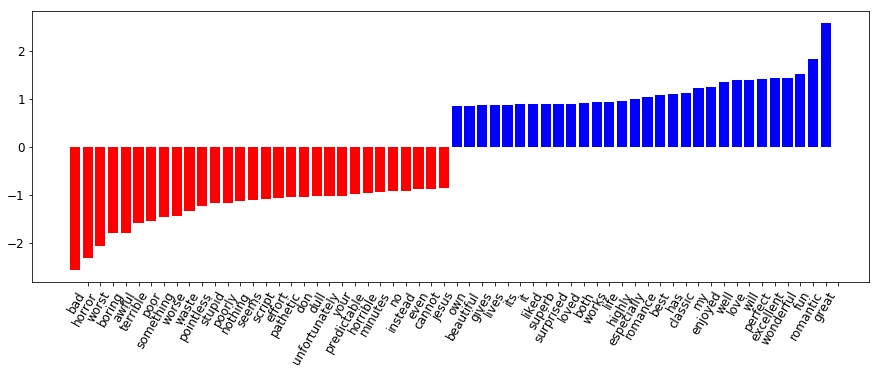}
         \caption{Trained on the original dataset}
         \label{fig:svm_orig_30}
     \end{subfigure}
     \vspace{10px}
      \begin{subfigure}[b]{\textwidth}
          \centering
          \includegraphics[width=0.8\linewidth]{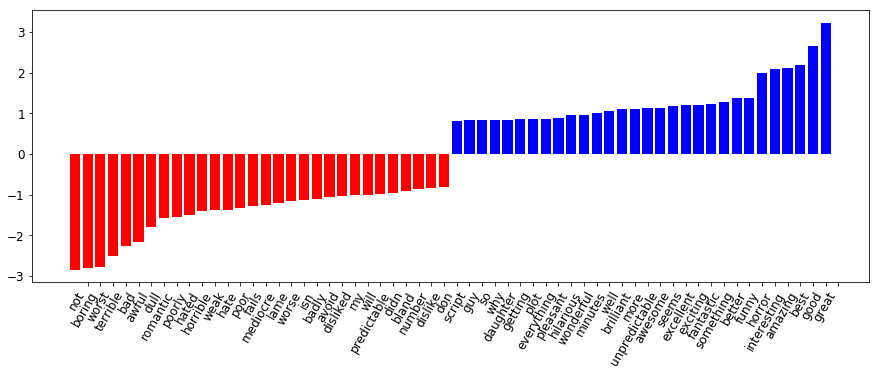}
          \caption{Trained on the revised dataset}
          \label{fig:svm_new_30}
      \end{subfigure}
      \vspace{10px}
     \begin{subfigure}[b]{\textwidth}
         \centering
         \includegraphics[width=0.8\linewidth]{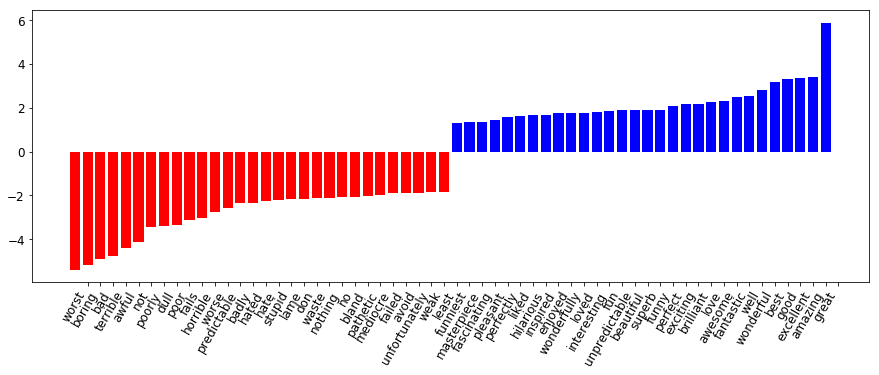}
         \caption{Trained on combined dataset}
         \label{fig:svm_augmented_30}
     \end{subfigure}
\caption{
Thirty most important features learned by an SVM classifier trained on TF-IDF bag of words.
\label{fig:svm_features_30}}
\end{figure}

\begin{figure}[t]
        \centering
        \fbox{\begin{subfigure}[t]{\textwidth}
            The blue box contains a text passage and a label. 
            Please edit this text in the textbox below, making a small number of changes such that:\vspace{2mm}
            
            (a) the document remains coherent and
            
            (b) the new label (colored) accurately describes the revised passage.\vspace{2mm}
            
            Do not change any portions of the passage unnecessarily.
            
            After modifying the passage and checking it over to make sure that is coherent and matches the label.
            \caption{Revising IMDb movie reviews\label{fig:sentiment_instructions}}
        \end{subfigure}}%
        \vspace{3mm}
        \fbox{\begin{subfigure}[t]{\textwidth}
            The upper blue box contains Sentence 1. The lower blue box contains Sentence 2.
            
            Given that Sentence 1 is True, Sentence 2 (by implication), must either be 
            
            (a) definitely True, (b) definitely False, or (c) May be True.\vspace{2mm}
            
            You are presented with an initial Sentence 1 and Sentence 2 and the correct initial relationship label (True, False, or May be True).\vspace{2mm}
            
            Please edit Sentence 2 in the textboxes, making a small number of changes such that:\vspace{2mm}
            
            (a) The new sentences are coherent and
            
            (b) The target labels (in red) accurately describe the truthfulness of the modified Sentence 2 given the original Sentence 1.\vspace{2mm}
            
            Do not change any portions of the sentence unnecessarily. 
            
            After modifying the text and checking it over to make sure that it is coherent and matches the target label.
            \caption{Revising hypothesis in SNLI\label{fig:hypothesis_instructions}}
        \end{subfigure}}
        
        \vspace{3mm}
        \fbox{\begin{subfigure}[t]{\textwidth}
            The upper blue box contains Sentence 1. The lower blue box contains Sentence 2.
            
            Given that Sentence 1 is True, Sentence 2 (by implication), must either be 
            
            (a) definitely True, (b) definitely False, or (c) May be True.\vspace{2mm}
            
            You are presented with an initial Sentence 1 and Sentence 2 and the correct initial relationship label (True, False, or May be True).\vspace{2mm}
            
            Please edit Sentence 1 in the textboxes, making a small number of changes such that:\vspace{2mm}
            
            (a) The new sentences are coherent and
            
            (b) The target labels (in red) accurately describe the truthfulness of the original Sentence 2 given the modified Sentence 1.\vspace{2mm}
            
            Do not change any portions of the sentence unnecessarily. 
            
            After modifying the text and checking it over to make sure that it is coherent and matches the target label.
            \caption{Revising premise in SNLI\label{fig:premise_instructions}}
        \end{subfigure}}
        \caption{Instructions used on Amazon Mechanical Turk for data collection}
    \end{figure} 

\end{document}